\documentclass[runningheads]{llncs}

 \usepackage{eccv}

\usepackage{array}
\usepackage{marvosym}
\usepackage{multirow}
\usepackage{pifont}
\usepackage{makecell} 
\usepackage{xcolor}
\definecolor{rowcolor}{RGB}{227,239,255}
\usepackage{tabularx}

\usepackage{tabularx}
\usepackage{booktabs}
\usepackage{colortbl}
\usepackage{xcolor}
\usepackage{soul}
\usepackage{amssymb}
\usepackage{multirow}
\usepackage{makecell}
\definecolor{upcolor}{RGB}{0, 205, 102}
\definecolor{downcolor}{RGB}{250, 128, 114}
\definecolor{rowcolor}{RGB}{227,239,255}
\usepackage{wrapfig}
\usepackage{eccvabbrv}
\usepackage{fontawesome5}
\usepackage{graphicx}
\usepackage{booktabs}
\usepackage[accsupp]{axessibility}  
\usepackage{wrapfig}
\usepackage[marginal]{footmisc}

\usepackage{hyperref}

\usepackage{orcidlink}

\begin{document}

\title{HART: High-Resolution Annotation-Free Reasoning Technique through a Closed-loop Framework} 

\titlerunning{HART}

\author{
Jiacheng Yang\inst{1}\thanks{~Equal contribution.}\and
Anqi Chen \inst{1*} \and
Yunkai Dang \inst{1} \and
Qi Fan \inst{1} \and
Cong Wang \inst{1,2} \and  \\
Wenbin Li  \inst{1,3}\thanks{~Corresponding authors.} \and
Feng Miao \inst{2\dag} \and
Yang Gao \inst{1}
}

\authorrunning{J.~Yang et al.}

\institute{
State Key Laboratory of Novel Software Technology, \\ Nanjing University, Nanjing, China \and
Institute of Brain-inspired Intelligence, Nanjing University, Nanjing, China \and
Shenzhen Research Institute of Nanjing University, Shenzhen, China
\email{liwenbin@nju.edu.cn, miao@nju.edu.cn}
}

\maketitle

\begin{abstract}
Current Large Multimodal Models (LMMs) struggle with high-resolution visual inputs during the reasoning process, as the number of image tokens increases quadratically with resolution, introducing substantial redundancy and irrelevant information.
A common practice is to identify key image regions and refer to their high-resolution counterparts during reasoning, typically trained with external visual supervision.
However, such visual supervision cues require costly grounding labels from human annotators.
Meanwhile, it remains an open question how to enhance a model’s grounding abilities to support reasoning without relying on additional annotations.
In this paper, we propose \textit{\textbf{H}igh-resolution \textbf{A}nnotation-free \textbf{R}easoning \textbf{T}echnique (HART)}, a closed-loop framework that enables LMMs to focus on and self-verify key regions of high-resolution visual inputs.
HART incorporates a post-training paradigm in which we design \textit{Advantage Preference Group Relative Policy Optimization (AP-GRPO)} to encourage accurate localization of key regions without external visual annotations.
Notably, HART provides explainable reasoning pathways and enables efficient optimization of localization.
Extensive experiments on MME-RealWorld-Lite, TreeBench, V* Bench, HR-Bench-4K/8K, and MMStar demonstrate that HART improves performance across a wide range of high-resolution visual tasks, consistently outperforming strong baselines.
Code will be available at \url{https://github.com/RL-MIND/HART}.


\end{abstract}

\section{Introduction}

Recent advances in Large Multimodal Models (LMMs) have attracted increasing attention from both industry and academia~\cite{dang2024explainable, hurst2024gpt, guo2025deepseek, bai2025qwen2, deepmind2025pro, jaech2024openai}.
LMMs have been widely applied to complex real-world tasks, such as object detection~\cite{liu2025visual, liu2024llava} and visual question answering~\cite{wang2024qwen2, wu2024deepseek, sun2024visual}.
They demonstrate strong capabilities in visual understanding and have the potential for enabling image-text interleaved reasoning.
Despite these advances, current LMMs still face a critical limitation: their performance degrades significantly on challenging high-resolution visual tasks~\cite{huang2025hires,zhan2025griffon,li2024monkey}.
In such tasks, the number of visual tokens increases significantly with the resolution of the input images, while only a small subset contains key information.
Popular LMM architectures, such as Qwen2.5-VL~\cite{bai2025qwen2} and InternVL3~\cite{zhu2025internvl3}, typically impose a maximum pixel constraint on input images to address this issue.
However, this constraint can lead to the loss of key information, restricting the model's visual perception capability~\cite{dong2024internlm,arif2025hired,DBLP:journals/corr/abs-2406-08487, liu2025oryx}.

\begin{figure}[tbp]
	\centerline{\includegraphics[width=0.835\columnwidth]{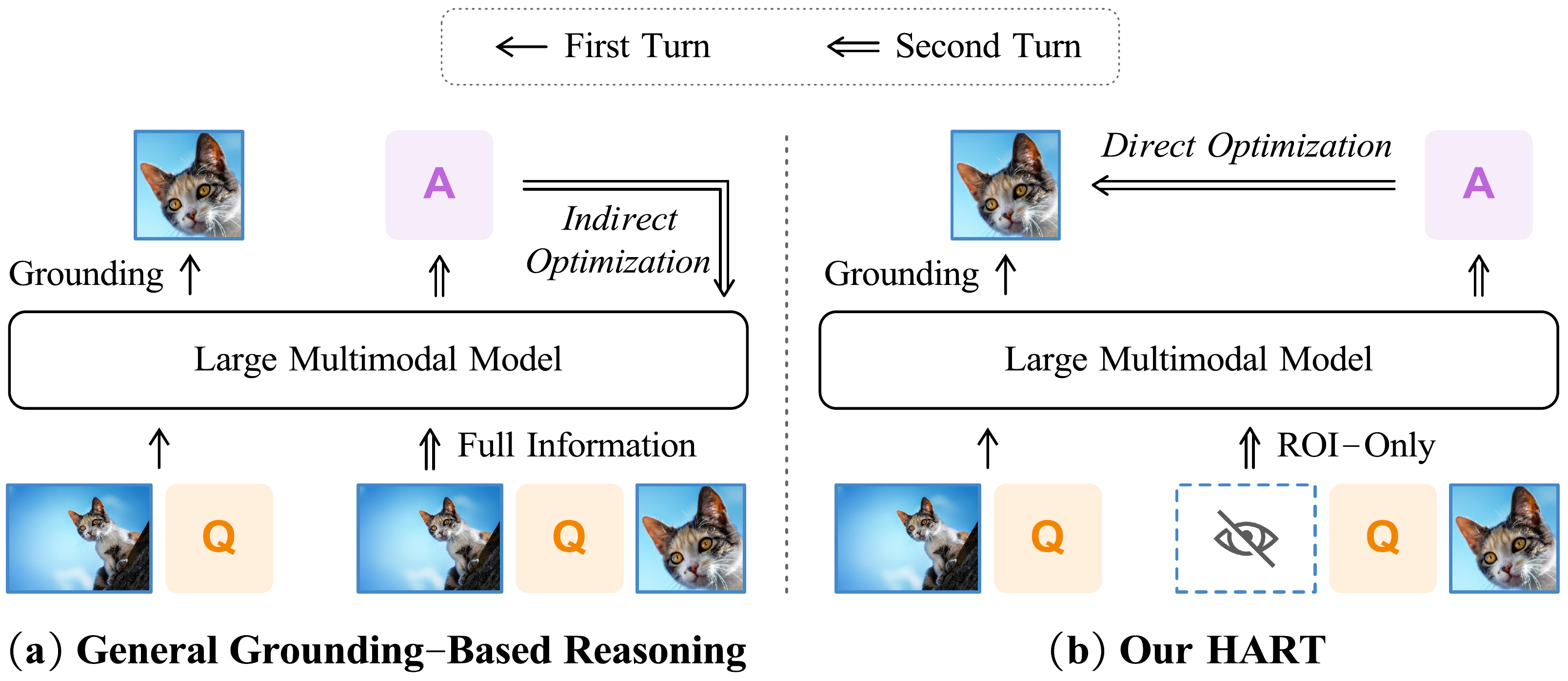}}
	\caption{Optimization procedures of (a) general grounding-based methods without bounding-box annotations and (b) our proposed model.
	General models indirectly optimize grounding performance, while HART performs direct optimization by answering based solely on the ROIs.
	Abbreviations: Q—Question; A—Answer.
	}
	\label{Procedures of (a) general unsupervised grounding-based models and (b) our proposed model.}

\end{figure}

To address this issue, existing works have explored a reasoning pathway that incorporates visual grounding, which is inspired by human visual processing~\cite{zhan2025griffon, DBLP:journals/corr/abs-2403-12966, DBLP:journals/corr/abs-2503-12605}.
Humans need to identify food and predators for survival in the wild, thus evolving the macula, a zone of acute vision in the retina~\cite{ptito2021retina}.
This structure guides visual attention and eye movements to locate key regions within high-resolution images.
Previous works have developed visual grounded reasoning models, attempting to equip LMMs with similar structures and functions~\cite{DBLP:journals/corr/abs-2504-18397, DBLP:journals/corr/abs-2404-09797}.
Conditioned on the textual question, they first predict key regions of interest (ROI) with the downsampled image and then solve the question based on both the downsampled image and the ROIs from the original resolution.
This pathway focuses only on critical visual information, thereby effectively reducing redundant computations.
There are two lines of research that study the optimization of visual grounded reasoning models.
On the one hand, some works directly enhance localization capabilities by using auxiliary visual annotations~\cite{shao2024visual, ni2025point, shen2025vlm}.
However, these methods require costly grounding labels from human annotators~\cite{wang2025traceable}.
On the other hand, recent research leverages reinforcement learning  (RL)~\cite{sutton1998reinforcement} to jointly optimize grounding and reasoning without relying on additional annotated data~\cite{su2025pixel, zheng2025deepeyes,huang2025high}.
These annotation-free approaches perform simple answer matching through a reward function.
The model's rewards measure the correctness of the final answer but cannot directly reflect localization accuracy.
Specifically, the model receives a positive reward when the answer is correct, even if the localization is incorrect.
Such reward misspecification will lead to negative optimization of grounding performance.
In early experiments, we found this issue to be relatively common, occurring in 36.5\% of cases for Qwen2.5-VL-7B~\cite{bai2025qwen2} and 63.8\% for InternVL3-8B~\cite{zhu2025internvl3}.
Therefore, it is natural to think:
\textbf{\textit{
How to directly optimize the grounding capabilities of LMMs without external visual annotations?
}}

To bridge this gap, this work aims to enable the model to self-verify its localization results for policy updates.
We propose a novel approach, \textit{\textbf{H}igh-resolution \textbf{A}nnotation-free \textbf{R}easoning \textbf{T}echnique (HART)}, to efficiently improve the performance of LMMs in high-resolution real-world scenarios without relying on extra visual annotations beyond final answers.
We design a closed-loop framework to overcome the limitation imposed by resolution constraints.
Given a textual question and a high-resolution image, our model identifies the ROIs and then crops relevant regions.
These regions serve as visual feedback guiding the reasoning process.
Subsequently, the original image is deliberately withheld, and the model answers the same question based solely on the cropped sub-images.
\cref{Procedures of (a) general unsupervised grounding-based models and (b) our proposed model.} compares the procedures of general annotation-free grounding-based methods and our proposed method.
Through this feedback loop, we introduce \textit{Advantage Preference Group Relative Policy Optimization (AP-GRPO)}, a reinforcement fine-tuning strategy that alleviates the reward misspecification problem and directly enhances grounding capabilities by introducing dynamic hyper-parameter adjustment.
Different from existing models~\cite{zheng2025deepeyes,huang2025high}, HART can directly optimize visual grounding, thereby further enhancing the model’s performance on perception-heavy tasks.
When applied to post-train Qwen2.5-VL-7B~\cite{bai2025qwen2}, HART achieves significant improvements across a range of challenging high-resolution benchmarks, \textit{i.e.}, $+20.1\%$ on MME-RealWorld~\cite{zhang2024mme}, $+6.7\%$ on TreeBench~\cite{wang2025traceable}, $+2.1\%$ on V* Bench~\cite{wu2024v}, and $+10.9\%$ on HR-Bench-8K~\cite{hrbench}.
The main contributions are summarized as follows:

\begin{itemize}
	\item We develop HART, a novel and interpretable framework that enhances the joint understanding of visual and textual inputs.
	It enables direct optimization of visual grounding without additional manual annotations.
	\item We introduce a reinforcement fine-tuning strategy, termed AP-GRPO, within the post-training paradigm to better incentivize the model to focus on key regions by prioritizing samples with correct grounding.
	\item We validate our method’s effectiveness on several high-resolution visual benchmarks and show that HART achieves state-of-the-art performance among methods supervised only by the final answer. 
\end{itemize}

\section{Related Work}

\subsection{Large Multimodal Models}

Recent breakthroughs in LMMs have significantly enhanced visual understanding capabilities, leading to growing popularity and widespread application across various domains~\cite{ li2024llava,DBLP:journals/corr/abs-2408-01800,liu2024improved, DBLP:journals/corr/abs-2406-08487, peng2025lmm, DBLP:journals/corr/abs-2503-12937, ye2024mplug}.
Inspired by DeepSeek-R1~\cite{guo2025deepseek}, many of these methods leverage reinforcement learning (RL) with reward engineering to shape effective thinking processes and solve increasingly complex reasoning tasks~\cite{wang2025vl, yang2025r1, meng2025mm,guo2025observe, zhan2025vision, thawakar2025llamav, tan2025reason, DBLP:journals/corr/abs-2503-12937}.
Despite recent progress, most LMMs still struggle with high-resolution image inputs and typically impose a resolution constraint due to limited visual-token capacity~\cite{liu2024hrvda, DBLP:journals/corr/abs-2403-04473, guo2024llava, liu2025oryx}.
This can result in blurred images and loss of key visual information, resulting in performance degradation.
Such limitations further restrict the application of LMMs in high-resolution real-world scenarios, such as remote sensing and autonomous driving~\cite{dong2024internlm}.
While recent works~\cite{zhang2025falcon ,DBLP:journals/corr/abs-2403-12895} attempt to address this issue, they often overlook the crucial role of visual grounding in multimodal reasoning.
In contrast, our work adopts a visual grounded reasoning process to address the challenges posed by high-resolution image analysis.

\subsection{Visual Grounded Reasoning Models}

Visual grounding LMMs aim to predict and focus on the key ROIs before actually answering the question~\cite{DBLP:journals/corr/abs-2503-12605}.
Existing studies highlight that this process is similar to human visual processing and can ensure more accurate answers through grounded reasoning~\cite{zhan2025griffon, DBLP:journals/corr/abs-2403-12966,DBLP:journals/corr/abs-2504-18397,DBLP:journals/corr/abs-2404-09797}.
There are two lines of work that study the optimization of visual grounding capabilities.
One direction focuses on directly fine-tuning models on labeled data that contain ground-truth bounding box annotations of key image regions~\cite{shao2024visual, ni2025point, shen2025vlm}.
The differences between the predicted and ground-truth bounding boxes are used to update the policies.
While such approaches can directly optimize the grounding capabilities of LMMs, they require large-scale and costly manual annotations~\cite{li2025self}.

Another line of research seeks to leverage end-to-end RL to improve grounding capabilities without external visual annotations~\cite{su2025pixel, zheng2025deepeyes, huang2025high}.
However, these RL-based approaches typically post-train the model leveraging a reward signal derived from only the correctness of the final answer~\cite{wang2025traceable}. 
Consequently, they do not directly optimize grounding performance.
We observed that annotation-free methods that directly optimize grounding capabilities remain undiscovered, as it is challenging to assess the grounding accuracy without bounding box annotations.
Leveraging visual feedback, we aim to directly optimize the grounding capability of LMMs.
Therefore, our work provides a more flexible solution that enables the model to self-verify its localization results for policy updates.

\section{Preliminaries}

\subsection{Grounding-Based Visual Reasoning }

Visual grounding refers to the task of localizing visual elements in an image based on a linguistic question. 
Grounding capabilities enable existing models to focus on critical visual regions that correspond to the given linguistic concepts, thereby effectively reducing redundant computations in high-resolution vision-centric tasks~\cite{DBLP:journals/corr/abs-2403-12966, DBLP:journals/corr/abs-2504-18397}.
Formally, given an input image $I_f$ and a textual question $q$, the visual grounding model first identifies the key sub-images $I_s$ that semantically align with the entities or relations described in $q$, denoted as $I_s \sim \pi_\theta(\cdot \mid I_f, q)$,
where $\pi_\theta$ denotes the model policy parameterized by $\theta$.
Next, the model leverages the task-relevant regions and generates an answer $a$ based on both the full image and key sub-images, that is, $a \sim \pi_\theta(\cdot \mid  I_f, I_s, q)$.

\subsection{Group Relative Policy Optimization}
Group Relative Policy Optimization (GRPO)~\cite{shao2024deepseekmath} is an efficient RL algorithm for post-training.
For a question-answer pair, GRPO first generates a group of $G$ candidate responses $\left\{ o_i\right\}^G_{i=1}$ and receives corresponding rewards $\left\{ r_i\right\}^G_{i=1}$. 
The advantage $A_i$ of each response is computed in a group-relative manner:
\begin{equation}
A_i = \frac{r_i - \text{mean}(\left\{ r_i\right\}^G_{i=1})}{\text{std}(\left\{ r_i\right\}^G_{i=1})},
\end{equation}
where $\text{mean}(\cdot)$ and $\text{std}(\cdot)$ are the mean and standard deviation of the rewards.
The policy $\pi_\theta$ is updated as follows: 
\begin{equation}
\mathcal{J}_{\text{GRPO}}(\theta) = \frac{1}{G} \sum^G_{i=1}(\frac{\pi_\theta(o_i \vert q)}{\pi_{\theta_{\text{old}}}(o_i \vert q)} A_i - \beta \mathbb{D}_{\text{KL}} (\pi_\theta \Vert \pi_{\text{ref}} )),
\end{equation}
where $\beta$ balances the KL-penalty term, $\pi_{\theta_{\text{old}}}$ is the old model, and $\pi_{\text{ref}}$ is the reference model.

\section{The Proposed Method}

In this section, we begin with pilot experiments designed to illustrate the challenges faced by annotation-free visual grounding methods, which serve as the research foundation for our study.
Then, we introduce the idea of our approach HART, a closed-loop framework that directly optimizes grounding and reasoning without relying on additional annotated data.

\subsection{Vanishing Advantages in Indirect Optimization}

\begin{figure*}[t]
	\centerline{\includegraphics[width=\textwidth]{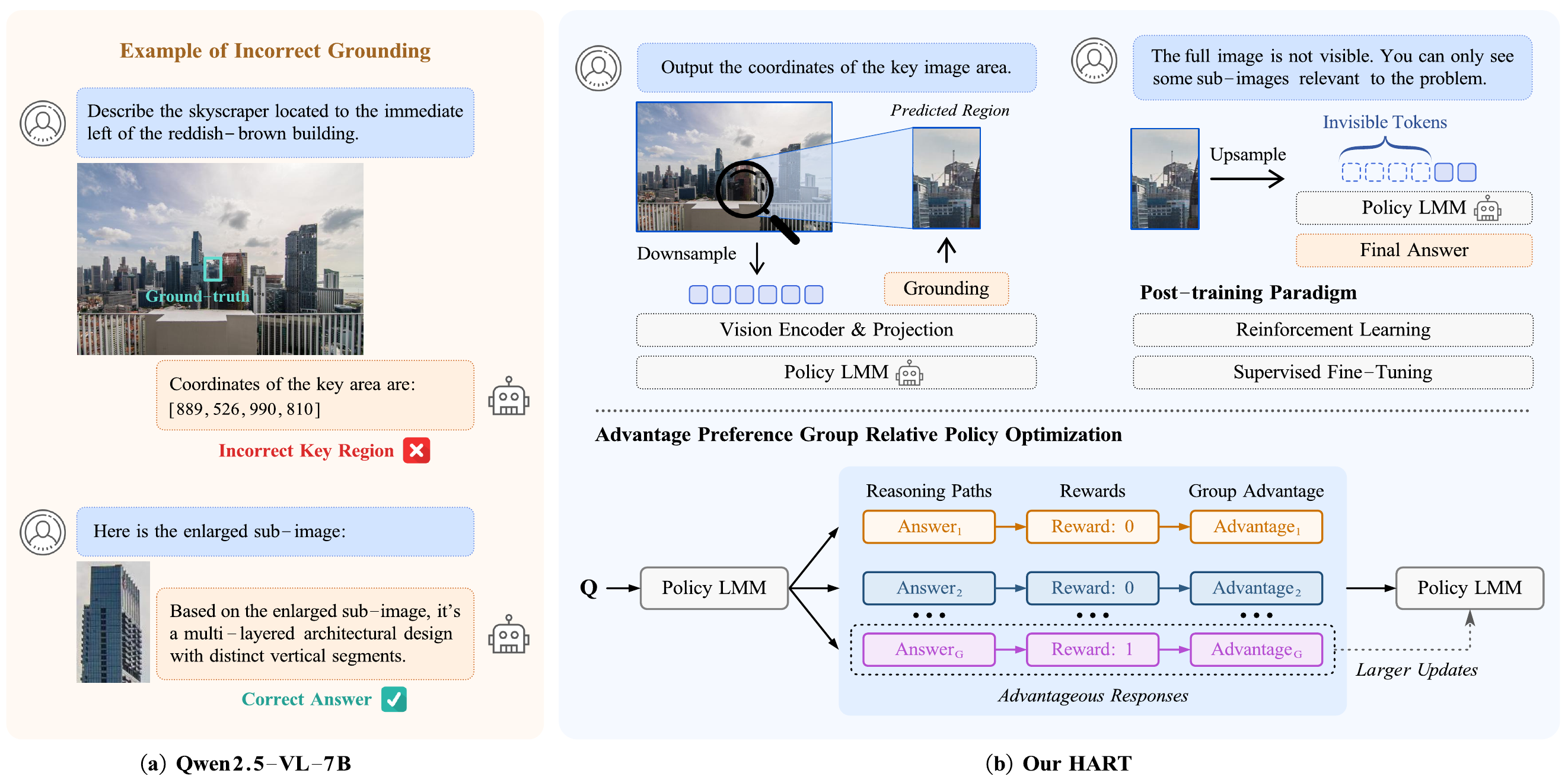}}
	\caption{Left: An example of Qwen2.5-VL-7B where the final answer is correct but the grounding is incorrect.
	Right: \textbf{HART Framework.}
	The post-training strategy consists of Reinforcement Learning (RL) and Supervised Fine-Tuning (SFT).
	In stage 1, after identifying the ROIs, the model answers based solely on the sub-regions and the original question.
	AP-GRPO is introduced to improve the model’s grounding capabilities.
	In stage 2, HART uses SFT to further enhance the high-resolution reasoning capabilities. 
	}
	\label{HART Framework.}

\end{figure*}

The rewards of annotation-free visual grounding methods measure the correctness of the final answer but cannot directly reflect grounding accuracy.
As a result, they receive a positive reward when the final answer is correct even though the grounding is incorrect, as illustrated in \cref{HART Framework.}(a).
This reward misspecification problem may lead to performance limitations.
While this phenomenon is theoretically possible, its real-world occurrence and frequency remain unclear. 
Therefore, we conduct pilot experiments on LMMs grounding and reasoning, taking Qwen2.5-VL-7B~\cite{bai2025qwen2} and InternVL3-8B~\cite{zhu2025internvl3} as representative examples.

\begin{table}[htp]
\caption{
The joint distribution of answer correctness and grounding correctness for Qwen2.5-VL-7B~\cite{bai2025qwen2} and InternVL3-8B~\cite{zhu2025internvl3} on the Visual CoT dataset~\cite{shao2024visual}. 
}
\label{The joint distribution of answer correctness and grounding correctness}
\tiny
\newcolumntype{C}{>{\centering\arraybackslash}X}
\newcolumntype{A}{>{\centering\arraybackslash}p{0.5cm}} 
\centering

\begin{tabularx}{\columnwidth}{p{1.8cm}c|CC|CC|c}
\toprule
\multirow{2}{*}{Method} &
\multirow{2}{*}{\textbf{HART (Ours)}} &
\multicolumn{2}{c|}{Correct answer} &
\multicolumn{2}{c|}{Incorrect answer} &
\multirow{2}{*}{\makecell[c]{Proportion of\\incorrect grounding\\given correct answer}}
\\
\cmidrule(lr){3-4}\cmidrule(lr){5-6}
& &
\makecell[c]{Incorrect\\grounding} &
\makecell[c]{Correct\\grounding} &
\makecell[c]{Incorrect\\grounding} &
\makecell[c]{Correct\\grounding} &
\\
\midrule

\multirow{2}{*}{Qwen2.5-VL-7B~\cite{bai2025qwen2}} &
\ding{55} & 1057 & 1838 & 466 & 681 & 36.5\% \\
 &  \ding{51} &  \textbf{502} & 1830 & 1028 & 682 & \textbf{21.5\%} \\
\midrule

\multirow{2}{*}{InternVL3-8B~\cite{zhu2025internvl3}} &
\ding{55} & 1359 & 770 & 1578 & 335 & 63.8\% \\
& \ding{51} & \textbf{998} & 788 & 1931 & 325 & \textbf{55.9\%} \\

\bottomrule
\end{tabularx}

\end{table}

\textbf{Experimental settings.} 
Both models are evaluated on the test set of the Visual CoT dataset~\cite{shao2024visual}, since it provides ground-truth bounding box annotations of key image regions.
Visual CoT contains a series of visual tasks, including text/doc, fine-grained understanding, charts, and relation reasoning.
We apply a filtering procedure to exclude (a) subsets that ask descriptive questions or have complex answer formats, to facilitate answer validation, and (b) questions with yes/no answers, to reduce the impact of random guessing on the statistics.
The resulting test set contains $4,042$ questions in total, each with one corresponding ground-truth bounding box.
We adopt intersection over ground-truth~\cite{YU2024110714,10887401} as the metric to evaluate grounding accuracy.
Specifically, grounding is considered correct if at least one predicted bounding box covers more than $0.3$ of the ground-truth area.

\textbf{Experimental results.} 
\cref{The joint distribution of answer correctness and grounding correctness} shows the distribution of the models’ responses by answer accuracy and grounding accuracy.
The responses are divided into four categories: correct answer with incorrect grounding, correct answer with correct grounding, incorrect answer with incorrect grounding, and incorrect answer with correct grounding.
Qwen2.5-VL-7B correctly answers $2,895$ questions on Visual CoT, of which $1,057$ bounding boxes are incorrectly localized.
Another popular model, InternVL3-8B, correctly answers $2,129$ questions on Visual CoT, of which $1,359$ bounding boxes are incorrectly localized.

\textbf{Conclusions of pilot experiments.}
Existing annotation-free visual grounding methods typically post-train LMMs using only the final answer as the training signal~\cite{zheng2025deepeyes,huang2025high}.
Based on our pilot experiments, we identify two limitations of these methods.
First, it becomes difficult to quantify the actual contribution of the grounding.
Second, in more than $36.5\%$ of the cases where the models receive a positive reward, the localization is incorrect.
Such reward misspecification can lead to negative optimization of grounding performance, as the model policy tends to encourage unreliable reasoning process. 
These limitations further restrict  performance on visually intensive tasks.

\subsection{HART}

To overcome the above limitations, a novel closed-loop framework named \textit{\textbf{H}igh-resolution \textbf{A}nnotation-free \textbf{R}easoning \textbf{T}echnique (HART)} is proposed to directly optimize high-resolution visual grounding and understanding.
As illustrated in \cref{HART Framework.}(b), we extend the idea of decomposing the reasoning process from current research~\cite{li2025self} to visual grounding LMMs.
We also take into account the maximum-token constraint adopted in many models and restrict image resolution in the first turn.
During the training phase, our proposed HART first prompts the model to predict the key ROIs conditioned on the down-sampled full input image $I_f$ and the textual question $q$.
The instruction is:
\textit{
Output the coordinates of the key image area relevant to the problem [$I_f, q$].}
The model’s response is returned in bounding-box format.
In the second step, HART assesses whether the visual grounding is sufficient for generating the final answer.
The ROIs $I_s$ are cropped from the original high-resolution image, and the original image is deliberately withheld.
The instruction is adjusted based on the visual feedback:
\textit{
You were supposed to answer a question based on a full image, but now the full image is not visible.
You can only see some sub-regions [$I_s$] relevant to the problem.
Answer the following question: [$q$].
}
As shown in \cref{The joint distribution of answer correctness and grounding correctness}, through the above procedure, the proportion of Qwen2.5-VL-7B~\cite{bai2025qwen2} and InternVL3-8B~\cite{zhu2025internvl3} producing a correct answer while localizing an incorrect region decreases by $15.0\%$ and $7.9\%$, respectively.
This result indicates that under the HART framework, the models’ responses more accurately reflect the reliability of localization.

Let $L \in \{0,1\}$ denote whether the model localizes the correct region and $R \in \{0,1\}$ denote whether it produces the correct response.
Compared to baseline pipelines~\cite{zheng2025deepeyes,huang2025high}, the probability of our model producing a correct answer while localizing the incorrect region is much lower, leading to a smaller entropy.
We state our result as follows. 
\begin{proposition} 
Let $I_{\text{HART}}(L;R)$ and $I_{\text{baseline}}(L;R)$ denote the mutual information between localization correctness $L$ and response correctness $R$ under our method and baselines, respectively. Then,
\begin{equation}
I_{\text {HART}}(L;R)>I_{\text{baseline}}(L;R).
\end{equation}
\end{proposition}
We prove this proposition in Appendix A.
The above expression shows that HART strengthens causal dependency between localization and reasoning. 
The feedback loop offers two benefits:
(a) It enables the model to self-verify its localization results, eliminating the need for manual bounding box annotations.
(b) The model can access more detailed visual information as we zoom in on the key region of the original image.
This design reduces redundant computations and overcomes maximum pixel constraints imposed by existing LMMs, thereby enhancing the effectiveness in high-resolution real-world scenarios.

\subsection{Advantage Preference Policy Optimization}

We propose a post-training strategy to further improve the model’s grounding and reasoning capabilities.
In the HART framework, a correct final response is more likely to indicate that the localized ROIs contain all the visual information needed to answer the question.
Hence, rewarding correct answers also encourages faithful grounding.
Motivated by this observation, we proceed to modify the standard GRPO algorithm~\cite{shao2024deepseekmath} to optimize the localization policy.

In vanilla GRPO, all samples are assigned equal weight. 
However, samples with incorrect responses clearly exhibit higher localization uncertainty, which can lead to reward misspecification within our feedback loop.
Therefore, different from the traditional post-training methods, we prefer correct and advantageous responses as they typically indicate faithful grounding.
Accordingly, we propose \textit{\textbf{A}dvantage \textbf{P}reference \textbf{G}roup \textbf{R}elative \textbf{P}olicy \textbf{O}ptimization (AP-GRPO)} to assign dynamic weights to responses based on their advantages, thus allowing the model to focus more on optimizing the samples with correct grounding.
In particular, given a textual question and the cropped sub-images, AP-GRPO first generates a group of candidate responses and receives corresponding rewards $\left\{ r_i\right\}^G_{i=1}$, where $r_i = 1$ indicates a correct answer and $r_i = 0$ indicates an incorrect answer.
Then the advantages of each response are computed, and the optimization objective is as follows:
\begin{gather}
\mathcal{J}_{\text{AP-GRPO}}(\theta) = \frac{{1}}{G} \sum^G_{i=1}({{\mu_1}}\frac{\pi_\theta(o_i \vert q)}{\pi_{\theta_{\text{old}}}(o_i \vert q)} A_i - {{\mu_2}} \mathbb{D}_{\text{KL}} (\pi_\theta \Vert \pi_{\text{ref}} )),\\
\mu_1 = 1 + k(r_i - \text{mean}(\left\{ r_i\right\}^G_{i=1})),\\
\mu_2 = \beta\,\Bigl(1 - k\bigl(r_i - \text{mean}(\{ r_i\}_{i=1}^G)\bigr)\Bigr),
\end{gather}
where the scaling factor $k$ is the only hyperparameter in HART.
First, the samples with correct grounding are assigned higher weights with the parameter $\mu_1$, which provides larger updates for advantageous responses.
Next, we introduce a dynamic weighting for the KL penalty coefficient.
The dynamic weight factor $\mu_2$ reduces the KL penalty when the grounding is correct, thereby allowing greater deviation from the reference model.

\textbf{Theoretical guarantees of AP-GRPO.} 
To theoretically  understand the advantages of the proposed AP-GRPO in comparison to  prior studies,  the following proposition verifies that  AP-GRPO reduces reward misspecification that typically  causes negative optimization of grounding performance.
The proof is provided in Appendix B.
\begin{proposition}
\label{policy gradient}
Let $g_{\text {AP-GRPO}}$ and $g_{\text{baseline}}$ denote the gradients of AP-GRPO and baselines, respectively.
Consider any single-step reinforcement learning environment with binary rewards. 
Then $\exists \alpha \in [0,1]$ such that
\begin{align}
\label{g_HART}
g_{\text {AP-GRPO}} = g_{\text{baseline}} - \alpha P(L=0, r=1)\mathbb{E}_{L=0, r=1}\left[ \nabla_\theta  \log \pi_\theta \right].
\end{align}
\end{proposition}
In other words, \cref{policy gradient} demonstrates that AP-GRPO effectively reduces the negative impact of reward misspecification seen in prior studies.
Consequently, answer correctness becomes a more accurate reflection of perception quality, which forms the theoretical foundation of the proposed AP-GRPO.
Compared with vanilla GRPO, which assigns equal weight to all samples, AP-GRPO has several significant advantages:
(a) It enhances visual understanding and feature extraction by encouraging attention to the correct ROIs within the image.
(b) The model’s grounding capabilities are directly optimized without relying on additional visual supervision, since AP-GRPO allows the reward signal to evaluate the grounding performance.
(c) The proposed strategy also ensures interpretable reasoning.

Although AP-GRPO enables the model to learn precise visual localization, withholding full visual information inevitably causes a decrease in answer accuracy.
Building upon the RL strategy, we further apply SFT to enhance the model’s high-resolution reasoning capabilities.
In the SFT phase, the original image is fully visible to the model. 
Following \cite{chen2025beyond}, we adopt a clean dataset separation:  $\mathcal{D}_{\text{SFT}}$ for SFT and $\mathcal{D}_{\text{RL}}$ for RL.
The loss function is represented as
\begin{equation}
\mathcal{L}(\theta) = - \mathbb{E}_{(x,y)\sim\mathcal{D_{\text{SFT}}}} \sum^T_{t=1} \text{log} P(y_t \vert x, y_{<t} ; \theta),
\end{equation}
where $T$ is the text length and $(x, y)$ represents the query and target response in dataset $\mathcal{D}_{\text{SFT}}$.

\section{Experiments}

\begin{table*}[tbp] 

    \caption{
    Answer accuracy of state-of-the-art models on the in-distribution dataset MME-RealWorld-Lite~\cite{zhang2024mme}. 
    \textbf{Bold} and \underline{underlined} indicate the best and second best results respectively.
    The numbers in parentheses indicate the number of samples for each sub-task.
    Abbreviations: 
    RS-Remote Sensing;
    MO-Monitoring; 
    DT-Diagram and Table;
    AD-Autonomous Driving; 
    OCR-Optical Character Recognition in the Wild.
    }
    
    \tiny  
    \newcolumntype{C}{>{\centering\arraybackslash}X}
    \centering
    \begin{tabularx}{\textwidth}{p{3.5cm}c|CCCCC|CCCC|c}
        \toprule 
        \multirow{2}{*}{Method}
        &\multirow{2}{*}{Parameters}
         & \multicolumn{5}{c|}{Perception} & \multicolumn{4}{c|}{Reasoning} &\multirow{2}{*}{\textbf{Overall}} \\  
         \cline{3-11}
        &  & AD (350) & MO (319) & OCR (250) & RS (150) & DT (100) & AD (400) & DT (100)& OCR (100) & MO (150) \\  
        
        \hline 
            \multicolumn{11}{l}{\textit{\textcolor{gray!115}{Open-source General Models}}} \\
        Qwen2.5-VL-7B~\cite{bai2025qwen2}
        & 7B & 30.0 & 27.3 & 87.6 & 32.7 & 83.0
        & 23.0 & 62.0 & 72.0 & 28.7 & 42.3 \\

        LLaVA-OneVision-7B~\cite{li2024llava}
        & 7B & 39.4 & 31.7 & 80.0 & 40.0 & 56.0
        & 32.0 & 33.0 & 65.0 & 38.0 & 43.7\\

        InternVL3-8B~\cite{zhu2025internvl3}
        & 8B & 36.9 & 34.5 & 83.6 & 49.3 & 75.0
        & 37.0 & 44.0 & 70.0 & 40.0 & 47.9 \\

        \hline
          \multicolumn{11}{l}{\textit{\textcolor{gray!115}{Qwen2.5-VL-7B with Post-Training}}} \\
          SFT
          & 7B & 43.7 & 43.6 & 85.6 & 55.3 & 78.0
          & 41.0 & 60.0 & 70.0 & 50.7 & 54.0\\

          GRPO~\cite{shao2024deepseekmath}
          & 7B & 43.7 & 42.9 & 81.6 & 51.3 & 75.0
          & 38.3 & 53.0 & 67.0 & 43.3 & 51.3\\
          
          GRPO + SFT
          & 7B & 49.7 & \underline{47.0} & 89.2 & 55.3 & \underline{87.0}
          & 39.0 & \underline{74.0} & \textbf{77.0} & \underline{60.0} & 58.1\\
          
          MGPO~\cite{huang2025high}
          & 7B & 44.0 & 46.7 & 86.4 & 54.0 & 78.0
          & 39.3 & 69.0 & 74.0 & 52.7 & 55.1\\
          
          MGPO + SFT
          & 7B & \underline{55.4} & 49.8 & 83.6 & \textbf{59.3} & 82.0
          & \underline{47.8} & 71.0 & 71.0 & \textbf{63.3} & \underline{60.5}\\
           \hline
         \multicolumn{11}{l}{\textit{\textcolor{gray!115}{Visual Grounded Reasoning Models}}} \\
         Pixel-Reasoner-7B~\cite{su2025pixel}
         & 7B & 30.9 & 38.9 & 89.6 & 52.0 & 86.0
         & 32.5 & 72.0 & 71.0 & 46.0 & 49.7 \\
         
         DeepEyes-7B~\cite{zheng2025deepeyes}
         & 7B & 33.4 & 43.3 & \textbf{90.0} & 52.7 & \textbf{89.0}
         & 35.0 & 69.0 & \underline{76.0} & 44.0 & 53.2\\
          \rowcolor{rowcolor}\textbf{HART-7B (Ours)} 
          & 7B & \textbf{57.7} & \textbf{49.8} & \underline{89.6} & \underline{58.7} & 86.0 
          & \textbf{51.0} & \textbf{75.0} & 72.0 & 58.7 & \textbf{62.4}\\ 
        \hline 
        \multicolumn{11}{l}{\textit{\textcolor{gray!115}{Larger-scale Open-source Models}}} \\
        Qwen2.5-VL-32B~\cite{bai2025qwen2} 
        & 32B & 40.7 & 29.5 & 87.2 & 40.7 & 83.0
        & 29.5 & 60.0 & 74.0 & 27.3 & 45.6 \\
        
        InternVL3-38B~\cite{zhu2025internvl3}
        & 38B & 40.0 & 42.6 & 85.6 & 56.0 & 71.0
        & 35.0 & 45.0 & 77.0 & 47.3 & 51.0 \\
        
        Qwen2.5-VL-72B~\cite{bai2025qwen2} 
        & 72B & 30.6 & 27.9 & {90.8} & 34.0 & 87.0
        & 25.5 & 61.0 & 74.0 & 26.7 & 43.7 \\
        
        LLaVA-OneVision-72B~\cite{li2024llava} 
        & 72B  & 40.0 & 37.9 & 79.2 & 50.7 & 67.0
        & 39.3 & 41.0 & 76.0 & 38.7 & 48.7 \\
        
        \hline 
        \multicolumn{11}{l}{\textit{\textcolor{gray!115}{Private Models}}} \\
        GPT-4o-mini~\cite{gpt4o-mini}
        & \textemdash
        & 24.2  & 26.5 & 62.5 & 6.7 & 44.2 
        & 26.8 & 39.1 & 47.0 & 25.8 & 36.4 \\
        
        Gemini-1.5-Pro~\cite{team2024gemini} 
        & \textemdash 
        & 26.6 & 31.1 & 67.6 & 14.0 & 39.9 
        & 19.2 & 33.2 & 52.7 & 28.3 & 38.2\\
        
        GPT-4o~\cite{hurst2024gpt}
        & \textemdash
        & 22.4 & 33.9 & 77.7 & 28.9 & 46.7 
        & 26.4 & 44.8 & 61.4 & 36.5 & 45.2 \\
        
        Claude 3.5 Sonnet~\cite{anthropic2024a}
        & \textemdash
        & 40.8 & 32.2 & 72.5 & 25.7 & 67.4
        & 31.9 & 61.2 & 61.9 & 41.8 & 51.6 \\
        
        \bottomrule
    \end{tabularx}  
    \label{Performance comparison with state-of-the-art methods on the in-distribution dataset MME-RealWorld-Lite.}

\end{table*}

\textbf{Benchmarks.} 
We evaluate the proposed method on several benchmarks targeting high-resolution visual understanding capabilities of LMMs. 
(a) The MME-RealWorld dataset~\cite{zhang2024mme} comprises challenging visual question-answering pairs, with an average resolution of $2,076 \times 1,434$.
We use its training set for post-training, randomly sampling 10K examples for $\mathcal{D}_{\text{RL}}$ and assigning the remainder to $\mathcal{D}_{\text{SFT}}$.
Its test set, also referred to as MME-RealWorld-Lite, contains $1,919$ samples and is used for in-distribution evaluation.
(b) TreeBench~\cite{wang2025traceable} serves as an out-of-distribution benchmark with an average image resolution of $2,152 \times 1,615$.
Specifically, TreeBench contains a total of $833$ manually annotated bounding boxes, providing a basis for evaluating grounding capabilities.
Each sample in the datasets follows a multiple-choice format.
The evaluation metric is the multiple-choice accuracy.

\textbf{Baselines.} 
We compare HART with several state-of-the-art baselines.
(a) Private models including GPT-4o~\cite{hurst2024gpt}, Gemini series~\cite{deepmind2025Flash}, etc.
(b) Open-source general models including Qwen2.5-VL series~\cite{bai2025qwen2}, LLaVA-OneVision series~\cite{li2024llava}, and InternVL3 series~\cite{zhu2025internvl3}.
(c) Representative visual grounded reasoning models including Pixel-Reasoner~\cite{su2025pixel} and DeepEyes~\cite{zheng2025deepeyes}.
(d) Post-training methods including GRPO~\cite{shao2024deepseekmath} and the recently proposed MGPO~\cite{huang2025high} for high-resolution vision-centric tasks.

\textbf{Implementation Details.}
We employ the Transformer Reinforcement Learning (TRL) framework~\cite{vonwerra2022trl} to enable distributed training and use Qwen2.5-VL-7B~\cite{bai2025qwen2} as the base model.
All reinforcement fine-tuning methods employ a binary reward function that evaluates answer correctness. 
For fair comparison, they are trained using the same training set of MME-RealWorld.
The hyperparameter $k$ in AP-GRPO is set to $0.6$.
Training details and cost are shown in Appendix C.

\subsection{Main Results}

We evaluate the answer accuracy of our proposed method HART on popular high-resolution visual benchmarks.
\cref{Performance comparison with state-of-the-art methods on the in-distribution dataset MME-RealWorld-Lite.} presents the accuracy comparison between HART and baselines on the in-distribution dataset.
HART achieves an accuracy of 62.4\% on MME-RealWorld-Lite, surpassing existing visual grounded reasoning models such as Pixel-Reasoner~\cite{su2025pixel} and DeepEyes~\cite{zheng2025deepeyes}. 
Notably, our HART outperforms the representative private and open-source models in most perception and reasoning tasks.
Compared to the base model Qwen2.5-VL-7B~\cite{bai2025qwen2}, the proposed method achieves remarkable improvements on challenging tasks, \textit{i.e.}, +26.0\% on Remote Sensing, +27.7\% on Perception-Autonomous Driving, and +30.0\% on Reasoning-Monitoring tasks, indicating enhanced fine-grained visual understanding.
We can observe that visual grounding methods are able to improve the model’s performance on high-resolution visual tasks.
Among them, HART achieves the highest average accuracy, demonstrating the effectiveness of our closed-loop framework.
Furthermore, we apply HART to post-train the larger foundation model Qwen2.5-VL-32B and the recent model Qwen3-VL-8B. 
Detailed results are provided in Appendix D.

\begin{table*}[tbp] 

    \caption{
    Answer accuracy of state-of-the-art models on the out-of-distribution dataset TreeBench~\cite{wang2025traceable}. 
    \textbf{Bold} and \underline{underlined} indicate the best and second best results respectively.
    The numbers in parentheses indicate the number of samples for each sub-task.
    }
    
    \tiny  
    \newcolumntype{C}{>{\centering\arraybackslash}X}
    \centering
    \begin{tabularx}{\textwidth}{p{3.4cm}c|CCCCC|CCCCC|c}
        \toprule 
        \multirow{2}{*}{Method}
        &\multirow{2}{*}{Parameters}
         &\rotatebox{75}{Attributes (29)} 
         & \rotatebox{75}{OCR (68)} 
         & \rotatebox{75}{Material (13)} 
         & \rotatebox{75}{Obj. Retr. (16)} 
         & \rotatebox{75}{Phy. State (23)} 
         & \rotatebox{75}{Comparison (44)} 
         & \rotatebox{75}{Ordering (57)} 
         & \rotatebox{75}{Con. \& Oc. (41)} 
         & \rotatebox{75}{Spa. Cont. (29)} 
         & \rotatebox{75}{Per. Trans. (85)}  
         &\multirow{2}{*}{\textbf{Overall}} \\  
         \cline{3-12}
        & & \multicolumn{5}{c|}{Perception} & \multicolumn{5}{c|}{Reasoning} & \\ 
        
        \hline 
        \multicolumn{11}{l}{\textit{\textcolor{gray!115}{Open-source General Models}}} \\
        Qwen2.5-VL-7B~\cite{bai2025qwen2} 
        & 7B  & 55.2 & 27.9 & 53.8 & \underline{62.5} & 56.5 & 43.2 & 35.1 & 39.0 & 44.8 & \underline{20.0} & 37.0\\
        LLaVA-OneVision-7B~\cite{li2024llava} 
        & 7B & 55.2 & 32.4 & 53.8 & 50.0 & 56.5 & 36.4 & 22.8 & 41.5 & \textbf{72.4} & 21.2 & 37.3\\
        InternVL3-8B~\cite{zhu2025internvl3} 
        & 8B  & 51.7 & 33.7 & \textbf{69.2} & 56.3 & 56.5 & 43.2 & 24.6 & 39.0 & 72.4 & \textbf{21.2} & 38.8\\

        \hline
        \multicolumn{11}{l}{\textit{\textcolor{gray!115}{Qwen2.5-VL-7B with Post-Training}}} \\

        GRPO~\cite{shao2024deepseekmath} 
        & 7B  & 44.8 & 51.5 & 46.2 & 50.0 & 56.5 
        & \underline{47.7} & 28.1 & 41.5 & 44.8 & 14.1 & 38.0\\
        
        GRPO + SFT
        & 7B  & 48.3 & 47.1 & 61.5 & 50.0 & 52.2
        & 43.2 &  28.1 & 43.9 & 55.2 & 15.3 & 38.5\\

        MGPO~\cite{huang2025high}
        & 7B & 51.7 & 48.5 & 61.5 & 68.8 & \underline{56.5}
        & 43.2 & 35.1 & 43.9 & 44.8 & 11.8 & 39.5 \\
        
        MGPO + SFT 
        & 7B & 58.6 & 48.5 & 61.5 & 56.3 & 52.2
        & 40.9 & \underline{31.6} & \textbf{51.2} & 51.7 & 14.1 & \underline{40.3} \\
        
         \hline
		
		\multicolumn{11}{l}{\textit{\textcolor{gray!115}{Visual Grounded Reasoning Models}}} \\
     	 DeepEyes-7B~\cite{zheng2025deepeyes} 
     	 & 7B  & 62.1 & \underline{51.5} & 53.8 & \textbf{68.8} & 65.2 & 47.7 & 24.6 & 36.6 & 51.7 & 11.8 & 37.5\\
     	 Pixel-Reasoner-7B~\cite{su2025pixel} 
    	  & 7B  & \underline{58.6} & 48.5 & 61.5 & 50.0 & \textbf{65.2} & 40.9 & 31.6 & 39.0 & 44.8 & 14.1 & 39.0 \\
		        
        \rowcolor{rowcolor}\textbf{HART-7B (Ours)} 
        & 7B & \textbf{62.1} & \textbf{55.9} & \underline{61.5} & 56.3 & 52.2 
        & \textbf{50.0} & \textbf{35.1} & \underline{48.8} & \underline{62.1} & 14.1 & \textbf{43.7}\\

        \hline
        \multicolumn{11}{l}{\textit{\textcolor{gray!115}{Larger-scale Open-source Models}}} \\
        Qwen2.5-VL-32B~\cite{bai2025qwen2} 
        & 32B  & 51.7 & 54.4 & 53.8 & 62.5 & 69.6 & 38.6 & 33.3 & 46.3 & 62.1 & 16.5 & 42.5\\

        InternVL3-38B~\cite{zhu2025internvl3} 
        & 38B & 51.7 & 51.5 & 61.5 & 68.8 & 52.2 & 38.6 & 33.3 & 56.1 & 65.5 & 12.9 & 42.0 \\

        Qwen2.5-VL-72B~\cite{bai2025qwen2} 
        & 72B  & 65.5 & 48.5 & 69.2 & 56.3 & 56.5 & 38.6 & 33.3 & 51.2 & 72.4 & 11.8 & 42.2\\

        LLaVA-OneVision-72B~\cite{li2024llava} 
        & 72B & 62.1 & 36.8 & 53.8 & 62.3 & 65.2 & 47.7 & 28.1 & 53.7 & 65.5 & 12.9 & 40.5 \\

        \hline 
        \multicolumn{11}{l}{\textit{\textcolor{gray!115}{Private Models}}} \\
        Gemini-2.5-Flash~\cite{deepmind2025Flash} 
        & \textemdash  & 48.3 & 75.0 & 53.9 & 68.8 & 69.6 & 43.2 & 19.3 & 56.1 & 72.4 & 15.3 & 45.9\\

        GPT-4o~\cite{hurst2024gpt} 
        & \textemdash  & 51.7 & 69.1 & 61.5 & 43.8 & 65.2 & 43.2 & 38.6 & 48.8 & 72.4 & 18.8 & 46.9\\

        Gemini-2.5-Pro~\cite{deepmind2025pro} 
        & \textemdash  & 51.7 & 83.8 & 61.5 & 75.0 & 56.5 & 54.6 & 36.8 & 65.9 & 86.2 & 20.0 & 54.1\\

        o3~\cite{openai2025o3} 
        & \textemdash & 69.0 & 79.4 & 69.2 & 68.8 & 65.2 & 50.0 & 38.6 & 61.0 & 86.2 & 22.4 & 54.8 \\
        
        \bottomrule
    \end{tabularx}  
    \label{Comparison with state-of-the-art methods on the out-of-distribution dataset TreeBench. }
\end{table*}

\begin{table}[tp] 

	\caption{Answer accuracy on other multimodal benchmarks. }
    
	\tiny  
		\newcolumntype{C}{>{\centering\arraybackslash}X}
		\centering
		\begin{tabularx}{\columnwidth}{p{1.8cm}lCCCC}
		\toprule 
		    Capability & Benchmark &  
            \makecell[c]{Qwen2.5-\\VL-7B~\cite{bai2025qwen2}} &  \makecell[c]{InternVL3-\\8B~\cite{zhu2025internvl3}}& 
            \makecell[c]{LLaVA-OneVision\\-7B~\cite{li2024llava}}&
            \makecell[c]{\textbf{HART-7B}\\{\textbf{(Ours)}}}\\
		\midrule
		\multirow{2}{=}{Multi-image VQA} 
		&BLINK \cite{fu2024blink} &55.2& 55.5&48.2& \textbf{56.8}\\
		 &Mantis-Eval \cite{jiang2024mantis} &70.8&70.1&64.2& \textbf{72.8}\\
		\midrule
		\multirow{2}{=}{Low-resolution VQA
		}&MathVista \cite{lu2024mathvista} &68.2&71.6&58.3& \textbf{71.8}\\
		&MMStar \cite{chen2024we} & 59.3 & 53.5 & 56.7 & \textbf{62.8}\\
        \midrule
		\multirow{3}{=}{High-resolution VQA}&V* Bench \cite{wu2024v}  & 78.5 & 72.3 & 70.7 & \textbf{80.6}\\
	    &HR-Bench-4K \cite{hrbench} &  70.1 & 70.8 & 64.3 & \textbf{71.1}\\
	    &HR-Bench-8K \cite{hrbench} & 61.0 & 62.0 & 59.8 & \textbf{71.9}\\
  		\bottomrule
    \end{tabularx}   
    \label{Answer accuracy on other multimodal benchmarks.}

\end{table}

Next, we evaluate the models on out-of-distribution datasets.
As shown in \cref{Comparison with state-of-the-art methods on the out-of-distribution dataset TreeBench. }, our HART also achieves open-source state-of-the-art on TreeBench with an accuracy of 43.7\%.
HART delivers significant improvements over the base model Qwen2.5-VL-7B~\cite{bai2025qwen2}, surpassing other post-training paradigms such as GRPO~\cite{shao2024deepseekmath} and MGPO~\cite{huang2025high}.
This demonstrates the effectiveness of the proposed post-training strategy in improving the joint understanding of visual and textual inputs.
In \cref{Answer accuracy on other multimodal benchmarks.}, we compare HART with the base model Qwen2.5-VL-7B on a range of multimodal benchmarks covering both low- and high-resolution visual tasks.
Detailed results are provided in Appendix D.
The results highlight the strong adaptability of our method across visual scenes of different resolutions.
Furthermore, we apply HART to post-train an additional base model InternVL3-8B~\cite{zhu2025internvl3}.
As shown in \cref{Answer accuracy compared with Qwen2.5-VL-7B}, HART achieves better performance, indicating robust training behavior.
In conclusion, our HART provides a flexible solution that adapts better to high-resolution real-world scenarios.

\begin{table}[tp]
\caption{
Answer accuracy compared with Qwen2.5-VL-7B~\cite{bai2025qwen2} and InternVL3-8B~\cite{zhu2025internvl3}.
}
\label{Answer accuracy compared with Qwen2.5-VL-7B}
\tiny
\newcolumntype{C}{>{\centering\arraybackslash}X}
\newcolumntype{A}{>{\centering\arraybackslash}p{0.5cm}} 
\centering

\begin{tabularx}{1\columnwidth}{p{1.8cm}c|CC|CC}
\toprule
\multirow{2}{*}{Method} &
\multirow{2}{*}{\textbf{HART (Ours)}} &
\multicolumn{2}{c|}{MME-RealWorld-Lite~\cite{zhang2024mme}} &
\multicolumn{2}{c}{TreeBench~\cite{wang2025traceable}}
\\
\cmidrule(lr){3-4}\cmidrule(lr){5-6}
& &
Perception &
Reasoning &
Perception &
Reasoning 
\\
\midrule

\multirow{2}{*}{Qwen2.5-VL-7B~\cite{bai2025qwen2}} &
\ding{55} &46.4 &35.9 &43.6 &33.2  \\
 &  \ding{51}  &\textbf{64.9} & \textbf{58.5} &\textbf{57.0} &\textbf{35.9}\\
\midrule

\multirow{2}{*}{InternVL3-8B~\cite{zhu2025internvl3}} &
\ding{55}  &51.1 & 42.9 & 46.3 &34.4 \\
& \ding{51} &\textbf{61.8} & \textbf{54.8} & \textbf{59.1} & \textbf{35.9}\\

\bottomrule
\end{tabularx}
\end{table}

\begin{table}[tp] 

	\caption{Grounding performance of AP-GRPO and baselines on the TreeBench~\cite{wang2025traceable} and Visual CoT~\cite{shao2024visual} datasets. 
	Bold numbers are the best results.}
    
	\tiny   
		\newcolumntype{C}{>{\centering\arraybackslash}X}
		\centering
		\begin{tabularx}{1\columnwidth}{p{3.5cm}|CC|CC}
		\toprule 
		   \multirow{2}{*}{ Method} & \multicolumn{2}{c|}{ TreeBench~\cite{wang2025traceable}} & \multicolumn{2}{c}{ Visual-CoT~\cite{shao2024visual}} \\
		   \cmidrule{2-5}
		    &  Incorrect ($\downarrow$) &  Correct ($\uparrow$)&  Incorrect ($\downarrow$) &  Correct ($\uparrow$)\\
		   \midrule
		Qwen2.5-VL-7B~\cite{bai2025qwen2} & 49.8\% & 50.2\% & 34.0\% & 66.0\% \\
		LLaVA-OneVision-7B~\cite{li2024llava} & 61.7\%  & 38.3\%  & 33.0\% & 67.0\% \\
		InternVL3-8B~\cite{zhu2025internvl3} & 84.9\%  & 15.1\%  & 71.1\%  & 28.9\% \\
		\hline
	    \multicolumn{5}{l}{\textit{Qwen2.5-VL-7B with Post-Training}} \\
	    \hline
        GRPO~\cite{shao2024deepseekmath} & 49.0\%  & 51.0\%  & 33.9\%  & 66.1\%  \\
	    MGPO~\cite{huang2025high} & 48.7\%  & 51.3\% & 33.7\%  & 66.3\% \\
	    \rowcolor{rowcolor}\textbf{AP-GRPO (Ours)} & \textbf{24.6\%}  & \textbf{75.4\%} & \textbf{22.3\%} & \textbf{77.7\% } \\
  		\bottomrule
    \end{tabularx}   
    
    \label{Grounding performance of HART and baselines on the out-of-distribution dataset TreeBench. }

\end{table}

\begin{figure}[bp]

	\centerline{\includegraphics[width=1\columnwidth]{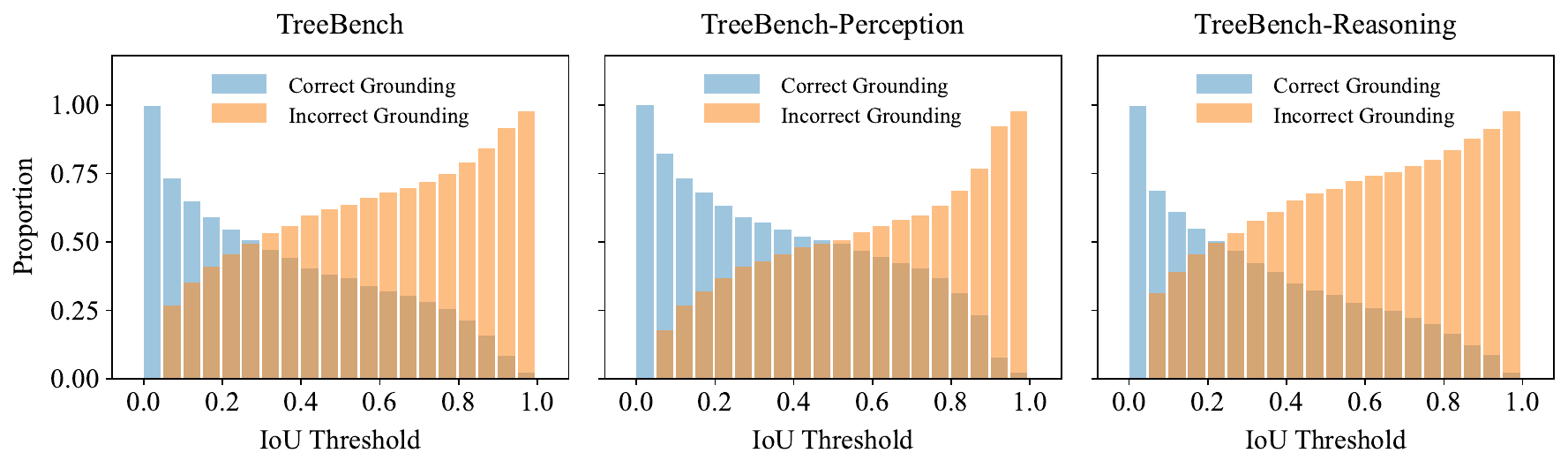}}
	\caption{Grounding performance of AP-GRPO on TreeBench~\cite{wang2025traceable}.
	}
	\label{Grounding performance of HART on TreeBench-Perception and TreeBench-Reasoning.}
\end{figure}

\subsection{Grounding Results}
In this section, we evaluate the grounding accuracy of the proposed method on the TreeBench~\cite{wang2025traceable} and Visual CoT~\cite{shao2024visual} datasets, both of which provide ground-truth bounding box annotations for key image regions relevant to the questions.
We utilize intersection over ground-truth as the metric to evaluate grounding accuracy.
\cref{Grounding performance of HART and baselines on the out-of-distribution dataset TreeBench. } presents the grounding performance with a coverage threshold of 0.3.
AP-GRPO achieves superior grounding capabilities on both benchmarks, indicating its effectiveness in accurately localizing key regions.
Compared to the base model Qwen2.5-VL-7B~\cite{bai2025qwen2}, the proposed method yields a $+25.2\%$ improvement on TreeBench~\cite{wang2025traceable} and a $+11.7\%$ improvement on Visual CoT~\cite{shao2024visual}.
While both AP-GRPO and MGPO adopt a visual grounded reasoning pipeline for post-training, the issue of reward misspecification limits MGPO’s ability to deliver substantial improvements over the base model Qwen2.5-VL-7B.
In contrast, HART can self-verify its localization results for policy updates, leading to more accurate localization.
We report the grounding performance of AP-GRPO over a range of Intersection over Union (IoU) thresholds (0.0 to 0.95) in \cref{Grounding performance of HART on TreeBench-Perception and TreeBench-Reasoning.}.
These results suggest that AP-GRPO offers an effective optimization strategy for enhancing both grounding and reasoning capabilities.
We evaluate grounding performance across different training stages and under more stringent intersection-over-ground-truth thresholds. 
Detailed results are shown in Appendix D.

\begin{table}[tbp] 

	\caption{Ablations of each component of our HART.}
    
	\tiny  
		\newcolumntype{C}{>{\centering\arraybackslash}X}
		\centering
		\begin{tabularx}{\columnwidth}{p{3.5cm}c|CC|CC}
		\toprule 
		   \multirow{2}{*}{Method} & \multirow{2}{*}{$k$}&  \multicolumn{2}{c|}{ MME-RealWorld-Lite~\cite{zhang2024mme}} & \multicolumn{2}{c}{ TreeBench~\cite{wang2025traceable}} \\
		   \cmidrule{3-6}
		    & &  Perception  &  Reasoning &  Perception & Reasoning \\
		   \midrule
		Qwen2.5-VL-7B~\cite{bai2025qwen2} & - & 46.4 & 35.9 & 43.6 & 33.2\\
		\hline
	    \multicolumn{5}{l}{\textit{+ Post-Training}} \\
	    \hline
	    GRPO~\cite{shao2024deepseekmath} & - & 52.1 & 49.1 & 52.3 & 32.0\\
	    MGPO~\cite{huang2025high} & - & 58.0 & 50.5 &  53.7 & 31.2\\
	    \textbf{HART-7B (Ours)}\\
        {  ·  {AP-GRPO}}  & $0.15$ & 51.2 & 47.5 & 51.0 & 32.4\\
	    {  ·  {AP-GRPO + SFT}}  & $0.15$ & \textbf{67.0} & 56.9 & 56.4 & 31.6\\
        {  ·  {AP-GRPO}}  & $0.30$ & 53.6 & 49.9 &55.0 & 33.6\\
	    {  ·  {AP-GRPO + SFT}}  & $0.30$ & 64.1 & 56.0 & \textbf{57.7} & 34.8\\
        {  ·  {AP-GRPO}}  & $0.60$ & 53.3 & 48.0 & 52.3 & 33.6\\
        { ·  {SFT + AP-GRPO}}  & $0.60$ & 51.7 & 48.1 & 55.0 & 32.4 \\
	    { ·  {AP-GRPO + SFT}}  & $0.60$ & 64.9 & \textbf{58.5} & 57.0 & \textbf{35.9} \\ 
  		\bottomrule
    \end{tabularx}   
    
    \label{Ablations of each component of our HART.}

\end{table}

\subsection{Ablation Studies}

In this experiment, we conduct ablation studies to investigate the influence of each component of HART.
As demonstrated in \cref{Ablations of each component of our HART.}, we compare the answer accuracy of post-trained Qwen2.5-VL-7B~\cite{bai2025qwen2} with GRPO~\cite{shao2024deepseekmath}, MGPO~\cite{huang2025high}, and the proposed HART on MME-RealWorld-Lite~\cite{zhang2024mme} and TreeBench~\cite{wang2025traceable}, respectively.
The results indicate that both RL (Stage 1) and SFT (Stage 2) are crucial for enhancing visual understanding, as HART yields considerably higher accuracy than either stage alone.
In the table we also present the results of the sensitivity analysis on hyperparameter $k$.
HART consistently improves performance under different values of $k$, demonstrating its robustness.
Detailed results are provided in Appendix D.

\begin{figure*}[t]
	\centerline{\includegraphics[width=\textwidth]{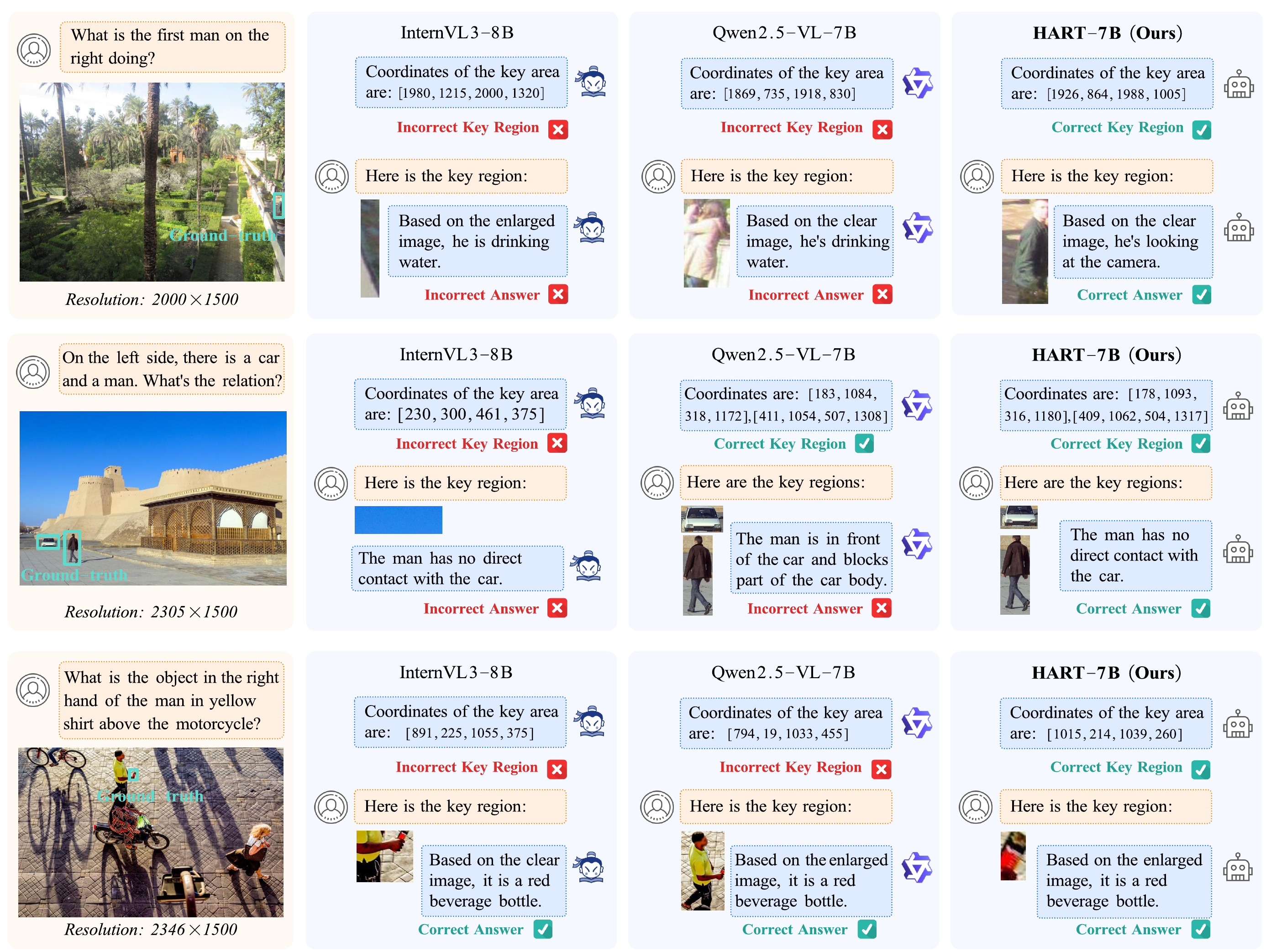}}
	\caption{Visualization of model outputs from InternVL3-8B~\cite{zhu2025internvl3}, Qwen2.5-VL-7B~\cite{bai2025qwen2}, and our method HART-7B on TreeBench~\cite{wang2025traceable}.
	}
	\label{Visualizations of the HART reasoning process.}

\end{figure*}

\subsection{Visualization Results}

We present a visualized comparison of our method against prior methods in \cref{Visualizations of the HART reasoning process.}. 
In this example, the task requires attending to the right-most man before executing the reasoning step. 
Compared with InternVL3-8B~\cite{zhu2025internvl3} and Qwen2.5-VL-7B~\cite{bai2025qwen2}, our method identifies the critical region more reliably and achieves a more accurate reasoning outcome.
More failure cases and visualization results are shown in Appendices E and F.

\section{Conclusion}

This paper proposes HART, a closed-loop framework designed for high-resolution visual tasks.
HART enables LMMs to identify and self-verify key regions of interest conditioned on visual and textual inputs.
To guide its localization behavior, we propose AP-GRPO to prioritize optimizing samples with correct grounding behavior, without relying on external visual supervision.
Empirical results demonstrate enhanced visual understanding across multiple high-resolution vision-centric tasks.
In summary, HART provides a foundation for further exploration in the joint optimization of grounding and reasoning capabilities.



\newpage

\section*{Acknowledgements}
This work is supported in part by the National Natural Science Foundation of China (62576160), Guangdong Basic and Applied Basic Research Foundation (2024A1515011340), the Fundamental Research Funds for the Central Universities (KG202508, KG202514), and 111 Center (B26023).

\bibliographystyle{splncs04}
\bibliography{main}
\end{document}